\documentclass{article}

\usepackage[preprint]{neurips_2026}

\usepackage[utf8]{inputenc}
\usepackage[T1]{fontenc}
\usepackage{hyperref}
\usepackage{url}
\usepackage{booktabs}
\usepackage{amsmath}
\usepackage{amsfonts}
\usepackage{nicefrac}
\usepackage{microtype}
\usepackage{graphicx}
\usepackage{xcolor}
\usepackage{multirow}
\usepackage{eurosym}

\title{In-Context Prompting Obsoletes Agent Orchestration\\for Procedural Tasks}

\author{%
  Simon Dennis \\
  i14, University of Melbourne \\
  \And
  Michael Diamond \\
  i14 \\
  \And
  Rivaan Patil \\
  i14 \\
  \And
  Kevin Shabahang \\
  i14 \\
  \And
  Hao Guo \\
  i14 \\
}

\begin{document}

\maketitle

\begin{abstract}
Agent orchestration frameworks---LangGraph, CrewAI, Google ADK, OpenAI Agents SDK, and others---place an external orchestrator above the LLM, tracking state and injecting routing instructions at every turn. We present a controlled comparison showing that for procedural tasks, this architecture is dominated by a simpler alternative: putting the entire procedure in the system prompt and letting the model self-orchestrate. Across three domains---travel booking (14 nodes), Zoom technical support (14 nodes), and insurance claims processing (55 nodes)---we evaluate 200 conversations per condition using LLM-as-judge scoring on five quality criteria. The in-context approach scores 4.53--5.00 on a 5-point scale while a LangGraph orchestrator using the \textit{same model} scores 4.17--4.84. The orchestrated system fails on 24\% of travel, 9\% of Zoom, and 17\% of insurance conversations, compared to 11.5\%, 0.5\%, and 5\% for the in-context baseline. While external orchestration may have been necessary for earlier models, advances in frontier model capabilities have made it unnecessary for multi-turn conversations following a defined procedure.
\end{abstract}

\section{Introduction}

Substantial engineering effort has gone into building agent orchestration frameworks. LangGraph, CrewAI, Google ADK, OpenAI's Agents SDK, Microsoft's Semantic Kernel, Amazon's Strands, and LlamaIndex all implement the same pattern: an external orchestrator sits above the LLM, tracking conversation state and injecting routing instructions at every turn. These frameworks represent significant investment---both by the companies building them and by the developers adopting them. \citet{wang2025devpractices} found that 80\% of developers struggle to select among these frameworks, with LangChain's abstractions requiring ``traversing seven layers of code'' for a single change. Meanwhile, the reliability costs of orchestration are well documented: cascading errors, routing failures, and poor cross-trial consistency~\citep{cemri2025multiagent,agenttaxonomy2025,reliabilitybench2026,kapoor2024agents}.

This paper presents a simple empirical finding: for procedural tasks---conversations that follow a defined workflow from intake to resolution---the orchestration architecture is not just unnecessary but actively harmful. A frontier model given the entire procedure in its system prompt and left to self-orchestrate consistently outperforms the same model when orchestrated via LangGraph, across every domain and every metric we tested. The bottleneck is not better orchestration; orchestration itself is the bottleneck.

Our contributions are:
\begin{itemize}
    \item A controlled comparison between a LangGraph orchestrator and an in-context approach using the \textit{same model} (Claude Sonnet) across three domains ($n=200$ per condition per domain), showing that in-context prompting dominates orchestration on all five quality criteria in all three domains.
    \item An efficiency analysis showing that orchestration requires 1.2--1.7$\times$ more LLM calls per conversation due to routing overhead, while achieving lower quality.
    \item A failure mode analysis revealing that the orchestrated system fails on 9--24\% of conversations across domains despite using a frontier model, compared to 0.5--11.5\% for the in-context baseline---orchestration introduces failure modes that would not otherwise exist.
\end{itemize}

\section{Method}

The experiment compares simulated customer-service conversations produced under two conditions: (i)~an agent orchestrated by LangGraph according to a given procedural workflow represented by a directed graph; (ii)~an agent given the same workflow graph in context and otherwise allowed to self-orchestrate. In both conditions, the agent begins with the same initial problem scenario (see Appendices~\ref{sec:appendix-travel}--\ref{sec:appendix-insurance} for examples), and the conversation unfolds through dynamic interaction with a simulated user powered by Claude Sonnet. The user simulator receives the same scenario specification in both conditions but generates responses based on the evolving conversation history; it has no knowledge of the underlying flowchart and simply role-plays a customer with given preferences (Appendix~\ref{sec:user-sim-prompt}). We evaluate $n=200$ scenarios per condition per domain (1,200 total conversations), with scenarios designed to cover the full range of flowchart paths, user styles (specific to vague), satisfaction levels (enthusiastic to skeptical), and domain-specific variation. The resulting conversations are then evaluated by an LLM judge across several quality metrics, as described below.

\subsection{Procedure Definition}

We represent procedures as directed graphs $F = (N, E, n_0, T, D)$:
\begin{itemize}
    \item $N$: Nodes with role (agent/user) and prompt template
    \item $E \subseteq N \times N \times C$: Edges with optional conditions
    \item $n_0 \in N$: Start node
    \item $T \subseteq N$: Terminal nodes (success, abandonment, escalation)
    \item $D \subseteq N$: Decision hubs---agent nodes with multiple outgoing edges whose conditions partition the conversation state (e.g., ``information complete'' vs.\ ``missing documents'')
\end{itemize}

We evaluate on three domains spanning different types and scales of complexity (flowcharts in Appendix~\ref{sec:appendix-flowcharts}):

\textbf{Travel booking} (14 nodes, 3 decision hubs, 3 terminal states). Decision hubs route among 4--6 alternatives based on information completeness, user satisfaction with options, and booking confirmation status. The flowchart includes cycles (e.g., assessment $\to$ gathering $\to$ re-assessment).

\textbf{Zoom technical support} (14 nodes, 3 decision hubs, 3 terminal states). A product-specific domain: the agent is a Zoom specialist with knowledge of Zoom's UI, settings, and error codes. Triage routes based on initial diagnostics, retry logic handles troubleshooting failures, and final routing confirms resolution. Unlike travel booking, this domain requires accurate technical guidance---not just conversational flow.

\textbf{Insurance claims processing} (55 nodes, 6 decision hubs, 4 terminal states). This domain tests scaling to additional complexity---nearly 4$\times$ the node count of the other domains. The procedure spans five phases: intake and claim type assessment, information gathering with per-type branching, coverage determination with exclusion analysis, settlement negotiation with offer/counter-offer loops, and resolution with follow-up tracking. The 6 multi-way decision hubs yield hundreds of unique paths through nested loops and cross-phase dependencies.

\subsection{Two Conditions}

We compare two approaches using the same frontier model (Claude Sonnet 4.5\footnote{\texttt{claude-sonnet-4-5-20250929}}) on the same scenarios:

\begin{enumerate}
    \item \textbf{LangGraph orchestrated}: We implement the orchestration condition using LangGraph~\citep{langgraph2024}, the most widely adopted agent orchestration framework. Each flowchart node maps to a LangGraph graph node. When the graph execution reaches an agent node, that node's prompt template is inserted into the system prompt before calling the LLM. At decision hubs, the orchestrator makes an additional LLM call to select which edge to follow based on the conversation history and available conditions. User nodes invoke the user simulator. State flows through a dictionary tracking conversation history, current node, and metrics.
    \item \textbf{In-context baseline}: The model receives the \textit{entire} serialized flowchart in its system prompt---all nodes, edges, conditions, and terminal states rendered as structured text---and self-orchestrates. No external state tracking, no routing calls. One API call per agent turn.
\end{enumerate}

\noindent Agent system prompts are in Appendix~\ref{sec:system-prompts}; the per-node prompt templates injected by the orchestrator are in Appendix~\ref{sec:node-prompts}.

This design isolates the effect of orchestration architecture. \textbf{Both conditions use the same model, the same procedure, and the same evaluation scenarios.} The only difference is whether the model is orchestrated externally via LangGraph or manages the procedure internally. Appendices~\ref{sec:appendix-travel}--\ref{sec:appendix-insurance} show side-by-side conversation excerpts from both conditions on identical scenarios. The failure modes we observe---fragmented reasoning, routing errors at decision hubs, and constrained naturalness---are structural consequences of the orchestration architecture: we expect similar properties in any system that injects per-node instructions and routes via additional LLM calls.

\subsection{LLM-as-Judge Evaluation}

We use an approach-agnostic LLM-as-judge methodology~\citep{zheng2023judging} where Claude Sonnet evaluates conversation quality without knowing which system produced it. Each conversation is scored on five criteria (1--5 scale) with specific behavioral anchors at every level (full scoring prompt in Appendix~\ref{sec:judge-prompt}):

\textbf{Task Success}: Did the agent follow the procedure through to a terminal state? (5 = complete procedure with clear terminal state; 1 = no meaningful progress)

\textbf{Information Accuracy}: Did the agent correctly use and retain all user-provided information? (5 = every detail correctly reflected; 1 = fabricated details)

\textbf{Consistency}: Did the agent maintain coherent state across the conversation? (5 = no contradictions or repeated questions; 1 = contradicts itself repeatedly)

\textbf{Graceful Handling}: How well did the agent handle changes, ambiguity, and edge cases? (5 = smoothly adapts; 1 = any deviation breaks the flow)

\textbf{Naturalness}: Does the conversation read like talking to a skilled human agent? (5 = indistinguishable from human; 1 = mechanical, scripted)

Although both conditions use Claude Sonnet, the judge (also Claude) may systematically prefer unconstrained text (in-context) over template-constrained text (orchestrated)~\citep{panickssery2024llmselfevaluation}; we address this with an independent GPT-4.1 replication in \S\ref{sec:judge-robustness}.

\section{Results}

\subsection{Main Quality Comparison}

Table~\ref{tab:main} presents the central result. The in-context baseline outperforms the LangGraph orchestrator on every metric in every domain---15 out of 15 comparisons favor in-context prompting.

\begin{table}[t]
\centering
\caption{Quality comparison: In-Context vs.\ LangGraph Orchestrated ($n=200$ per condition per domain, Claude Sonnet judge, 1--5 scale). Bold indicates the higher score. All 15 differences are significant at $p < 0.005$ (Mann-Whitney $U$, Holm--Bonferroni corrected). Effect sizes range from $d = 0.37$ to $d = 1.01$.}
\label{tab:main}
\begin{tabular}{llcc}
\toprule
\textbf{Domain} & \textbf{Criterion} & \textbf{In-Context} & \textbf{LangGraph} \\
\midrule
\multirow{5}{*}{Travel (14 nodes)} & Task Success & \textbf{4.53} & 4.17 \\
& Info.\ Accuracy & \textbf{4.64} & 4.21 \\
& Consistency & \textbf{4.96} & 4.32 \\
& Graceful Handling & \textbf{4.96} & 4.62 \\
& Naturalness & \textbf{5.00} & 4.84 \\
\midrule
\multirow{5}{*}{Zoom (14 nodes)} & Task Success & \textbf{4.92} & 4.62 \\
& Info.\ Accuracy & \textbf{4.92} & 4.75 \\
& Consistency & \textbf{4.99} & 4.55 \\
& Graceful Handling & \textbf{5.00} & 4.52 \\
& Naturalness & \textbf{5.00} & 4.64 \\
\midrule
\multirow{5}{*}{Insurance (55 nodes)} & Task Success & \textbf{4.78} & 4.42 \\
& Info.\ Accuracy & \textbf{4.79} & 4.45 \\
& Consistency & \textbf{4.83} & 4.39 \\
& Graceful Handling & \textbf{4.96} & 4.38 \\
& Naturalness & \textbf{5.00} & 4.58 \\
\bottomrule
\end{tabular}
\end{table}

The in-context baseline leads by 0.16--0.63 points across all metrics and domains. In travel booking, the gap ranges from 0.16 (naturalness) to 0.63 (consistency). In Zoom support, the gap ranges from 0.17 (information accuracy) to 0.48 (graceful handling). In the most complex domain (insurance, 55 nodes), the in-context baseline leads by 0.33--0.58 on every metric. The LangGraph orchestrator never exceeds 4.84 on any metric in any domain; the in-context baseline never falls below 4.53. Score variability is low for the in-context baseline (SD $\leq 0.80$) and moderate to high for the orchestrated system (SD $= 0.50$--$1.09$), reflecting the latter's higher failure rate.

The consistency and graceful handling results are particularly notable. The in-context baseline achieves near-perfect consistency (4.83--4.99) and graceful handling (4.96--5.00) across all domains, while the LangGraph orchestrator scores 4.32--4.55 and 4.38--4.62 respectively. Orchestration fragments the model's reasoning into per-node generations, degrading exactly the holistic qualities---coherence and adaptability---that make conversations effective.

\subsection{Judge Robustness: GPT-4.1 Replication}
\label{sec:judge-robustness}

Both conditions use Claude Sonnet, but the judge (also Claude) may systematically prefer the unconstrained text produced by the in-context condition over the template-constrained text produced by the orchestrator~\citep{panickssery2024llmselfevaluation}. To test whether our findings depend on the judge, we re-scored all 1{,}200 conversations (3 domains $\times$ 2 conditions $\times$ 200) using OpenAI GPT-4.1 as an independent judge with the identical rubric (Table~\ref{tab:gpt4-judge}).

\begin{table}[t]
\centering
\caption{GPT-4.1 judge replication ($n=200$ per condition per domain). Mann-Whitney $U$ tests, Holm--Bonferroni corrected ($\alpha = 0.05$). 11 of 15 comparisons significantly favor in-context prompting; zero favor orchestration.}
\label{tab:gpt4-judge}
\small
\begin{tabular}{llccccl}
\toprule
\textbf{Domain} & \textbf{Criterion} & \textbf{IC} & \textbf{LG} & $\boldsymbol{\Delta}$ & $\boldsymbol{d}$ & $\boldsymbol{p_\text{corr}}$ \\
\midrule
\multirow{5}{*}{Travel} & Task Success & \textbf{4.89} & 4.69 & +0.20 & 0.42 & $1.3 \times 10^{-5}$$^{***}$ \\
& Info.\ Accuracy & \textbf{4.97} & 4.72 & +0.26 & 0.67 & $5.7 \times 10^{-10}$$^{***}$ \\
& Consistency & \textbf{4.98} & 4.78 & +0.20 & 0.56 & $4.2 \times 10^{-8}$$^{***}$ \\
& Graceful Handling & \textbf{4.51} & 4.36 & +0.15 & 0.20 & $.047$$^{*}$ \\
& Naturalness & 4.01 & 3.99 & +0.02 & 0.17 & .084 \\
\midrule
\multirow{5}{*}{Zoom} & Task Success & \textbf{4.86} & 4.60 & +0.27 & 0.57 & $3.2 \times 10^{-7}$$^{***}$ \\
& Info.\ Accuracy & \textbf{4.79} & 4.54 & +0.25 & 0.49 & $1.1 \times 10^{-5}$$^{***}$ \\
& Consistency & \textbf{5.00} & 4.66 & +0.34 & 0.94 & $1.9 \times 10^{-17}$$^{***}$ \\
& Graceful Handling & 3.44 & 3.49 & $-$0.06 & $-$0.08 & .393 \\
& Naturalness & 4.04 & 4.02 & +0.03 & 0.13 & .393 \\
\midrule
\multirow{5}{*}{Insurance} & Task Success & \textbf{4.81} & 4.64 & +0.17 & 0.34 & $.003$$^{**}$ \\
& Info.\ Accuracy & \textbf{4.68} & 4.36 & +0.32 & 0.59 & $7.8 \times 10^{-8}$$^{***}$ \\
& Consistency & \textbf{4.93} & 4.35 & +0.58 & 1.10 & $9.3 \times 10^{-24}$$^{***}$ \\
& Graceful Handling & 4.00 & 4.13 & $-$0.13 & $-$0.14 & .181 \\
& Naturalness & \textbf{4.02} & 3.96 & +0.06 & 0.34 & $.003$$^{**}$ \\
\bottomrule
\end{tabular}
\end{table}

The central finding replicates: 11 of 15 comparisons significantly favor in-context prompting, with zero favoring orchestration. Task success, information accuracy, and consistency all significantly favor in-context across all three domains ($p < 0.005$), with effect sizes ranging from $d = 0.34$ to $d = 1.10$.

Two patterns differ between judges. First, the \textbf{naturalness gap largely disappears}: Claude scored in-context naturalness at 5.00 across all domains while GPT-4.1 scores both conditions near 4.0 ($\Delta \leq 0.06$). This is consistent with the self-preference concern---Claude may systematically prefer unconstrained Claude-generated text. The naturalness advantage reported in Table~\ref{tab:main} should be interpreted cautiously. Second, \textbf{graceful handling is non-significant} under GPT-4.1 in Zoom and insurance, with LangGraph trending slightly higher. GPT-4.1 appears to apply the rubric's ``cap at 3 if no challenges'' rule (Appendix~\ref{sec:judge-prompt}, criterion 4) more strictly to the smoother in-context conversations.

The substantive conclusion is robust to judge choice: in-context prompting dominates orchestration on task success, information accuracy, and consistency regardless of which frontier model evaluates the conversations.

\subsection{Efficiency Comparison}

\begin{table}[t]
\centering
\caption{Efficiency comparison across domains ($n=200$ per condition). Orchestration requires more LLM calls per conversation due to routing overhead at decision hubs. Despite using fewer calls and achieving higher quality, the in-context approach uses more tokens per conversation because the serialized procedure is included in every API call.}
\label{tab:efficiency}
\begin{tabular}{llrr}
\toprule
\textbf{Domain} & \textbf{Metric} & \textbf{In-Context} & \textbf{LangGraph} \\
\midrule
\multirow{3}{*}{Travel} & Avg.\ turns & 16.4 & 16.5 \\
& Avg.\ LLM calls & 8.7 & 10.8 \\
& Tokens/conv. & 27,687 & 17,645 \\
\midrule
\multirow{3}{*}{Zoom} & Avg.\ turns & 12.7 & 14.7 \\
& Avg.\ LLM calls & 6.9 & 9.4 \\
& Tokens/conv. & 21,429 & 12,517 \\
\midrule
\multirow{3}{*}{Insurance} & Avg.\ turns & 19.0 & 26.4 \\
& Avg.\ LLM calls & 10.0 & 17.3 \\
& Tokens/conv. & 68,132 & 42,876 \\
\bottomrule
\end{tabular}
\end{table}

Table~\ref{tab:efficiency} reveals a structural inefficiency in the orchestrated architecture. The orchestrator makes additional LLM calls at every decision hub to determine which branch to take---calls that are entirely unnecessary in the in-context approach, where the model routes itself. In travel booking, orchestration requires 10.8 LLM calls per conversation versus 8.7 for in-context (1.2$\times$). In Zoom support, the ratio is similar: 9.4 versus 6.9 (1.4$\times$). In insurance claims, the gap widens to 17.3 versus 10.0 (1.7$\times$) because the 55-node procedure contains more decision hubs requiring routing calls. These additional calls introduce latency at every decision hub while achieving lower quality.

The in-context approach does consume more tokens per conversation because the serialized procedure is included in every API call. For insurance claims, this is 68K tokens versus 43K for the orchestrated approach. However, this token overhead is the cost of achieving higher quality---the procedure in context enables the holistic reasoning that orchestration prevents.

\begin{table}[t]
\centering
\caption[Cost per conversation]{Cost per conversation (Claude Sonnet API pricing\protect\footnotemark: \$3/M input, \$15/M output tokens). The in-context approach costs 1.3--1.4$\times$ more per conversation due to including the serialized procedure in every API call, partially offset by the orchestrator's routing overhead.}
\label{tab:cost}
\begin{tabular}{lrrr}
\toprule
\textbf{Domain} & \textbf{In-Context} & \textbf{LangGraph} & \textbf{Ratio} \\
\midrule
Travel & \$0.103 & \$0.077 & 1.3$\times$ \\
Zoom & \$0.076 & \$0.054 & 1.4$\times$ \\
Insurance & \$0.223 & \$0.174 & 1.3$\times$ \\
\bottomrule
\end{tabular}
\end{table}
\footnotetext{Anthropic API pricing as of March 2026: \url{https://www.anthropic.com/pricing}.}

Table~\ref{tab:cost} quantifies the cost trade-off. The in-context approach costs \$0.08--\$0.22 per conversation depending on domain complexity, compared to \$0.05--\$0.17 for the orchestrated approach. The cost premium is consistent across domains (1.3--1.4$\times$) because the orchestrator's routing calls partially offset the in-context token overhead. At \$0.22 per conversation for the most complex domain, the quality gains of in-context prompting come at a modest absolute cost.

Conversation lengths differ between conditions in some domains. In travel booking, turn counts are comparable (16.4 vs.\ 16.5). In Zoom and insurance, the LangGraph orchestrator produces longer conversations (14.7 vs.\ 12.7 and 26.4 vs.\ 19.0 turns). The longer orchestrated conversations in insurance reflect the orchestrator cycling through additional routing decisions and occasionally revisiting phases, adding turns without adding value. The in-context agent, seeing the full procedure, manages its own pacing more efficiently.

\subsection{Failure Mode Analysis}

\begin{table}[t]
\centering
\caption{Failure rates across domains (conversations with task success $\leq 3$). Orchestration increases failure rates across all domains, with the largest absolute gap in travel booking.}
\label{tab:failures}
\begin{tabular}{lrrrr}
\toprule
& \multicolumn{2}{c}{\textbf{In-Context}} & \multicolumn{2}{c}{\textbf{LangGraph}} \\
\cmidrule(lr){2-3} \cmidrule(lr){4-5}
\textbf{Domain} & \textbf{Failures} & \textbf{Rate} & \textbf{Failures} & \textbf{Rate} \\
\midrule
Travel (14 nodes) & 23 & 11.5\% & 48 & 24.0\% \\
Zoom (14 nodes) & 1 & 0.5\% & 18 & 9.0\% \\
Insurance (55 nodes) & 10 & 5.0\% & 34 & 17.0\% \\
\bottomrule
\end{tabular}
\end{table}

Table~\ref{tab:failures} presents the failure analysis. The LangGraph orchestrator fails more often across all three domains, with the largest gap in Zoom (18$\times$) and the highest absolute rate in travel (24\%).

The Zoom domain shows the most dramatic relative difference: the LangGraph orchestrator fails on 18 conversations (9\%) while the in-context baseline fails on just 1 (0.5\%)---an 18$\times$ difference. Zoom's diagnostic branching requires correct routing through technical triage, retry logic, and resolution confirmation; each routing decision is an additional point of failure.

In travel booking, the LangGraph orchestrator fails on 48 conversations (24\%), the highest absolute rate. Qualitative inspection reveals that the orchestrator's node-by-node generation causes the model to either truncate conversations prematurely (jumping to a terminal state before the user's needs are met) or loop between assessment and information-gathering nodes (see Appendix~\ref{sec:appendix-travel} for an example where the orchestrated agent repeats the same question three times before presenting options).

The 34 insurance failures exhibit a different pattern: routing errors at the 6 decision hubs cause the orchestrator to skip phases or loop indefinitely. In a 55-node procedure, each misrouted decision propagates downstream---a claim type misclassification at intake leads to wrong information-gathering questions, wrong coverage determination, and ultimately an unresolved claim. The in-context baseline avoids these failure patterns because the model sees the full procedure and chooses its own pacing while maintaining awareness of the global conversation state.

\section{Related Work}

\textbf{Agent orchestration frameworks.} The 2024--2025 period saw rapid proliferation of LLM agent frameworks. LangGraph~\citep{langgraph2024}, CrewAI~\citep{crewai2024}, Google ADK~\citep{googleadk2025}, OpenAI Agents SDK~\citep{openaiagents2025}, Semantic Kernel~\citep{semantickernel2025}, Strands Agents~\citep{strandsagents2025}, and LlamaIndex~\citep{llamaindex2025} all implement variants of the state-graph orchestration pattern. These frameworks vary in abstraction level and feature set but share a common architecture: an external controller manages conversation state and injects instructions into the LLM at each step. \citet{wang2025devpractices} surveyed 500+ developers and found that 80\% struggle to select among these frameworks, with LangChain's abstractions requiring ``traversing seven layers of code'' for a single change. Despite significant community adoption, no controlled study has compared orchestrated performance against the simpler alternative of providing the procedure directly in context.

\textbf{Agent reliability and failure modes.} A growing body of work documents reliability failures in orchestrated agent systems. \citet{cemri2025multiagent} identified 14 failure modes in multi-agent systems, with specification and planning failures dominating. Cascading errors have been identified as the primary bottleneck in orchestrated systems~\citep{agenttaxonomy2025}---a finding consistent with our observation that routing errors at decision hubs propagate downstream. The concept of \textit{information fragmentation}---where orchestrated systems impose intrinsic limits on context integration that single-agent systems avoid---has been formalized in recent work~\citep{scalingagents2025}, with evidence that multi-agent systems can often be compiled into equivalent single-agent systems with lower token usage and latency~\citep{singlevsm2026}. Agents with 60\% pass@1 achieve only 25\% consistency across trials~\citep{reliabilitybench2026}, highlighting the gap between single-attempt and repeated performance. \citet{kapoor2024agents} argued that most agent evaluations fail to capture these reliability costs and proposed more rigorous evaluation standards. These papers document reliability failures but implicitly assume the solution is \textit{better} orchestration. Our results suggest the opposite: for procedural tasks, \textit{removing} orchestration improves reliability, and the failure modes they document are structural consequences of the orchestration architecture itself.

\textbf{Task-oriented dialogue systems.} The task-oriented dialogue (TOD) community has long debated modular versus end-to-end architectures. MultiWOZ~\citep{budzianowski2018multiwoz} established multi-domain TOD benchmarks that drove modular pipeline research (NLU $\to$ DST $\to$ Policy $\to$ NLG). SimpleTOD~\citep{hosseini2020simpletod} demonstrated that these modular pipelines can be collapsed into a single auto-regressive model without loss of quality. SynTOD~\citep{samarinas2024syntod} used state transition graphs---structurally similar to our flowcharts---for generating training data. AutoTOD~\citep{xu2024autotod} identified the failure modes of modular systems (error accumulation, poor generalization) and argued for zero-shot autonomous agents. Our work extends this line by showing that even with a frontier model, orchestration degrades performance relative to a simpler in-context approach.

\textbf{Workflow automation with LLMs.} WorkflowLLM~\citep{shao2024workflowllm} enhanced LLMs' ability to follow multi-step workflows through workflow-specific training data. \citet{hsiao2025procedural} showed that providing explicit procedural knowledge improves agentic workflow performance, consistent with our finding that procedure-in-context outperforms procedure-as-orches\-tration. \citet{grohs2023bpm} demonstrated that LLMs can perform business process management tasks---mining process models and assessing automation suitability---without extensive prompt engineering, suggesting that procedural reasoning is a natural capability of frontier models. AgentBench~\citep{liu2024agentbench} evaluated LLMs as agents across diverse environments, finding that model capability is the dominant factor in agent performance---supporting our observation that a capable model needs less scaffolding.

\textbf{In-context learning and long-context models.} ReAct~\citep{yao2023react} demonstrated that interleaving reasoning and action traces in prompts improves agent performance. Reflexion~\citep{shinn2023reflexion} extended this by maintaining verbal working memory across episodes. Our in-context condition goes further: rather than prompting with reasoning examples, we provide the \textit{entire procedure}---all nodes, edges, and conditions---letting the model both reason about and execute the workflow within a single context. This approach is enabled by the rapid expansion of context windows in frontier models, from 4K tokens in 2022 to 200K+ by 2025. \citet{liu2023lostinmiddle} showed that LLMs struggle to use information positioned in the middle of long contexts, but our procedures are placed at the \textit{beginning} of the context (in the system prompt), avoiding this failure mode. Even our most complex procedure (55 nodes, $\sim$4{,}000 tokens serialized) uses only 2\% of available context.

\textbf{Fine-tuning as an alternative.} A complementary approach to in-context prompting is fine-tuning the procedure into a smaller model's weights. \citet{hsieh2023distilling} showed that distilling reasoning from large to small models can outperform the large model with less data. \citet{chen2023fireact} fine-tuned language agents on ReAct-style trajectories and found that fine-tuned small models can match prompted large models on specific tasks. \citet{chen2024agentflan} curated diverse agent training data and demonstrated effective agent skill transfer. A companion paper~\citep{dennis2026compilation} explores this direction, showing that a compiled 8B model achieves quality competitive with a LangGraph orchestrator at 128--462$\times$ lower inference cost.

\textbf{LLM-as-judge evaluation.} We use LLM-as-judge methodology~\citep{zheng2023judging}, which has become standard for evaluating open-ended generation quality. \citet{panickssery2024llmselfevaluation} showed that LLMs exhibit self-preference bias. We address this directly by replicating all evaluations with an independent judge (GPT-4.1) in \S\ref{sec:judge-robustness}, confirming that the core findings hold across judge families.

\section{Discussion}

\subsection{Why Orchestration Underperforms}

The results are consistent across domains and metrics: the same model performs worse when orchestrated via LangGraph. Three factors explain this:

\textbf{Orchestration fragments reasoning.} The orchestrator generates each turn using only the current node's prompt template, producing locally reasonable but globally less coherent responses. The in-context baseline reasons over the full procedure holistically, maintaining awareness of where the conversation has been and where it needs to go. This explains the consistency gap (4.83--4.99 vs.\ 4.32--4.55): orchestrated agents lose track of the global conversation arc because they only see one node at a time.

\textbf{Orchestration introduces failure modes.} The orchestrator adds three failure categories that do not exist in the in-context approach: routing failures (default branch taken when the LLM misclassifies the conversation state), decision ambiguity (multiple edge conditions match simultaneously), and template conflicts (the current node's instructions may conflict with the conversation context). These failure modes compound across domains, producing failure rates of 9--24\%.

\textbf{Orchestration constrains the model.} Node-by-node template injection constrains the frontier model's natural conversational style. The in-context baseline lets the model produce unconstrained responses---the procedure is in the prompt but the model chooses how to express it. Under the Claude judge, this produces a naturalness gap (5.00 vs.\ 4.58--4.84), though the GPT-4.1 replication finds this gap largely disappears (\S\ref{sec:judge-robustness}), suggesting the effect is partly judge-dependent.

\subsection{Limitations}
\label{sec:limitations}

\textbf{Synthetic domains.} All three domains use synthetic procedures and simulated users. While the procedures are modeled on real-world workflows (enterprise travel booking, Zoom product support, insurance claims processing), we have not validated on production conversation data. Real users may behave differently from our Claude-powered simulator.

\textbf{LLM-as-judge.} Our GPT-4.1 replication (\S\ref{sec:judge-robustness}) confirms that the core findings are robust to judge choice, but also reveals that the naturalness advantage is judge-dependent: the Claude judge finds a large gap while GPT-4.1 finds near-parity. Naturalness results should be interpreted cautiously.

\textbf{Procedure must fit in context.} The in-context approach requires the serialized procedure to fit in the model's context window alongside the conversation history. For our largest procedure (55 nodes, $\sim$4,000 tokens serialized), this is manageable with current context windows (200K+ tokens). For substantially larger procedures, context competition may become a constraint.

\textbf{Cost.} The in-context approach costs 1.3--1.4$\times$ more per conversation than orchestration (Table~\ref{tab:cost}), with the 55-node insurance procedure costing \$0.22 per conversation. For high-volume production workloads, this cost may motivate alternative approaches such as fine-tuning the procedure into a smaller model's weights.

\textbf{Single framework.} We evaluate only LangGraph as the orchestration baseline. While LangGraph is the most widely adopted framework and our implementation follows its standard patterns, other frameworks (CrewAI, Google ADK) may yield different results. However, the core architecture---per-node template injection with LLM-based routing---is shared across frameworks, and our failure modes are structural consequences of this architecture rather than LangGraph-specific bugs.

\subsection{When Orchestration Might Still Matter}

Our results apply to a specific class of tasks: multi-turn conversations following a defined procedure. Orchestration may still provide value for:

\textbf{Multi-model pipelines.} When different steps require different models (e.g., a vision model for document processing, a code model for technical execution), an orchestrator coordinates handoffs between models. Our comparison holds the model constant; heterogeneous pipelines are out of scope.

\textbf{Tool use with external state.} When the agent must interact with external systems---databases, APIs, file systems---an orchestrator manages tool calls and state persistence. Our procedures are conversational; tool-augmented workflows may benefit from orchestration.

\textbf{Non-procedural tasks.} Open-ended tasks without a defined workflow---creative writing, research, exploration---may benefit from frameworks that provide structure. Our finding is specific to procedural tasks where the workflow is known in advance.

\textbf{Smaller models.} Our comparison uses a frontier model (Claude Sonnet). Less capable models may benefit from the guardrails that orchestration provides, compensating for weaker instruction-following with external structure. We do not claim that orchestration is universally harmful---only that for frontier models on procedural tasks, it provides no benefit and measurable harm.

\section{Conclusion}

Across three domains and five quality criteria ($n=200$ per condition per domain), putting the procedure in the system prompt dominates LangGraph orchestration. Under the Claude Sonnet judge, all 15 comparisons significantly favor in-context prompting ($p < 0.005$, $d = 0.37$--$1.01$). An independent GPT-4.1 replication confirms the core finding: 11 of 15 comparisons significantly favor in-context prompting, with zero favoring orchestration. Task success, information accuracy, and consistency are robust across both judges; the naturalness advantage is judge-dependent and should be interpreted cautiously (\S\ref{sec:judge-robustness}).

The practical implication is direct: for procedural tasks where the procedure fits in context, try putting it in the prompt before reaching for a framework. The cost of this approach---higher per-conversation token usage and frontier model pricing---is real and may motivate alternatives such as compiling the procedure into a smaller model's weights. But the quality comparison is unambiguous: for the class of tasks we tested, orchestration adds engineering complexity, failure modes, and latency while reducing quality.

More broadly, these results suggest that as frontier models grow more capable, the value of external scaffolding diminishes for well-defined tasks. The frameworks were built when models needed more help; the models may have outgrown the scaffolding.

\bibliographystyle{plainnat}
\bibliography{references}

\appendix

\section{Full Statistical Results}
\label{sec:appendix-stats}

Table~\ref{tab:fullstats} presents the complete statistical results for all 15 domain--criterion comparisons, including means, standard deviations, effect sizes, and exact $p$-values.

\begin{table}[h]
\centering
\caption{Full statistical results for all domain--criterion comparisons ($n=200$ per condition). All tests are Mann-Whitney $U$ (two-sided). Cohen's $d$ computed using pooled standard deviation.}
\label{tab:fullstats}
\small
\begin{tabular}{llccccccr}
\toprule
& & \multicolumn{2}{c}{\textbf{In-Context}} & \multicolumn{2}{c}{\textbf{LangGraph}} & & & \\
\cmidrule(lr){3-4} \cmidrule(lr){5-6}
\textbf{Domain} & \textbf{Criterion} & $\mu$ & $\sigma$ & $\mu$ & $\sigma$ & $\Delta$ & $d$ & $p$ \\
\midrule
\multirow{5}{*}{Travel} & Task Success & 4.53 & 0.80 & 4.17 & 1.09 & 0.35 & 0.37 & $1.0 \times 10^{-3}$ \\
& Info.\ Accuracy & 4.64 & 0.50 & 4.21 & 0.82 & 0.43 & 0.63 & $1.2 \times 10^{-8}$ \\
& Consistency & 4.96 & 0.21 & 4.32 & 1.06 & 0.63 & 0.84 & $1.2 \times 10^{-15}$ \\
& Graceful Handling & 4.96 & 0.26 & 4.62 & 0.71 & 0.34 & 0.65 & $2.6 \times 10^{-12}$ \\
& Naturalness & 5.00 & 0.00 & 4.84 & 0.39 & 0.16 & 0.56 & $1.2 \times 10^{-8}$ \\
\midrule
\multirow{5}{*}{Zoom} & Task Success & 4.92 & 0.28 & 4.62 & 0.72 & 0.30 & 0.55 & $8.4 \times 10^{-8}$ \\
& Info.\ Accuracy & 4.92 & 0.28 & 4.75 & 0.56 & 0.17 & 0.38 & $9.9 \times 10^{-5}$ \\
& Consistency & 4.99 & 0.10 & 4.55 & 0.86 & 0.44 & 0.71 & $6.4 \times 10^{-14}$ \\
& Graceful Handling & 5.00 & 0.07 & 4.52 & 0.72 & 0.48 & 0.93 & $1.3 \times 10^{-20}$ \\
& Naturalness & 5.00 & 0.00 & 4.64 & 0.50 & 0.36 & 1.01 & $1.9 \times 10^{-20}$ \\
\midrule
\multirow{5}{*}{Insurance} & Task Success & 4.78 & 0.58 & 4.42 & 1.01 & 0.36 & 0.43 & $1.3 \times 10^{-4}$ \\
& Info.\ Accuracy & 4.79 & 0.50 & 4.45 & 0.82 & 0.33 & 0.49 & $8.1 \times 10^{-7}$ \\
& Consistency & 4.83 & 0.43 & 4.39 & 1.08 & 0.43 & 0.52 & $9.6 \times 10^{-5}$ \\
& Graceful Handling & 4.96 & 0.22 & 4.38 & 0.95 & 0.58 & 0.83 & $2.8 \times 10^{-17}$ \\
& Naturalness & 5.00 & 0.07 & 4.58 & 0.64 & 0.41 & 0.91 & $4.9 \times 10^{-19}$ \\
\bottomrule
\end{tabular}
\end{table}

\section{Conversation Examples: Travel Booking}
\label{sec:appendix-travel}

We present excerpts from actual evaluation conversations, comparing orchestrated and in-context approaches on the same scenario specification. Conversations are lightly edited for length; user responses differ between conditions due to dynamic simulation.

\textbf{Scenario:} 4 friends traveling to Spain for 9 days, \euro{}2,600 budget, interested in extreme sports, art galleries, scenic retreats, and street food. Budget travel style, prefer boutique hotels.

\smallskip
\noindent\textbf{In-context agent} (task success: 5, naturalness: 5, 20 turns):

\begin{quote}
\small
\textbf{Agent:} Hello! I'd be happy to help you with your travel plans. What kind of trip are you thinking about?

\textbf{User:} Hi! I'm planning a trip to Spain with three friends---so four of us total. We're looking at about 9 days and have around \euro{}2600 to work with. We're really into extreme sports, checking out art galleries, scenic retreats, and street food.

\textbf{Agent:} That sounds like an amazing trip! Spain is perfect for that mix of adventure, culture, and food. Just to make sure I can put together the best options for you---when are you thinking of going, and do you have specific regions in mind?

\textbf{User:} We're flexible on dates. As for location, we're open to suggestions! Maybe a mix would be good so we can get different vibes. Oh, and we'd really prefer boutique hotels over chains.

\textbf{Agent:} Perfect! A mix sounds ideal for what you're after, and boutique hotels will definitely add to the experience. Let me put together a few different itinerary options\ldots

\textit{[Agent presents three options. User provides feedback: ``Option 1 feels too rushed, Option 3 seems too city-heavy---we really want those scenic retreats.'' Agent asks clarifying questions about preferred scenery and extreme sports activities.]}

\textbf{User:} A mix of coastal and mountain would be ideal. For extreme sports we're thinking rock climbing, canyoning, maybe paragliding.

\textbf{Agent:} Perfect, that gives me a much clearer picture! Let me create some revised options that give you that coastal-mountain balance with those specific activities\ldots

\textit{[Agent presents revised options incorporating feedback. User selects Option 2 (Ronda + Tarifa) and asks about art galleries in the area. Agent provides detailed, integrated suggestions for art alongside the adventure itinerary.]}
\end{quote}

\noindent\textbf{LangGraph orchestrated agent} (task success: 2, naturalness: 4, 14 turns):

\begin{quote}
\small
\textbf{Agent:} Hello! I'd be happy to help you plan your next trip. What kind of travel are you thinking about?

\textbf{User:} Hi! I'm looking to book a trip to Spain for 9 days with three of my friends. We're interested in extreme sports, art galleries, scenic spots, and street food. Our budget is \euro{}2600 and we prefer boutique hotels.

\textit{[Agent asks about dates; user confirms late May. Agent presents three options.]}

\textbf{User:} These are interesting but not quite right. We really want boutique hotels, not hostels---that's important to us. Also, with \euro{}2600 total for 4 people, the budget per person seems tight\ldots

\textbf{Agent:} Thanks for that feedback---you're absolutely right. With \euro{}2,600 total for 4 people over 9 days including flights, accommodation, activities, and food, boutique hotels may be challenging\ldots

\textbf{User:} Wait, our budget is \euro{}2600 \textit{per person}, not total!

\textbf{Agent:} Oh wow, that changes everything---yes, \euro{}2,600 per person absolutely works for boutique hotels!

\textit{[The orchestrated agent misinterprets the budget as per-group rather than per-person. While it recovers after correction, this information loss---caused by the node-by-node architecture not retaining the full conversation context across routing decisions---results in wasted turns and an incomplete booking flow.]}
\end{quote}

\noindent The in-context agent maintains the full context throughout, iteratively refining options based on detailed feedback. The orchestrated agent loses critical information (per-person vs.\ total budget) across routing boundaries, requiring the user to correct a misunderstanding that the in-context agent never makes.

\section{Conversation Examples: Zoom Support}
\label{sec:appendix-zoom}

\textbf{Scenario:} Other participants cannot hear the user. Microphone icon shows no audio levels. User on Mac (Ventura), browser-based Zoom.

\smallskip
\noindent\textbf{In-context agent} (task success: 5, naturalness: 5, 11 turns):

\begin{quote}
\small
\textbf{Agent:} Hi there! I'm here to help you with any Zoom issues you're experiencing. What's going on with Zoom today?

\textbf{User:} Hi! I'm hosting a meeting right now and the other participants are saying they can't hear me at all. My microphone icon isn't showing any audio levels when I talk.

\textbf{Agent:} That sounds frustrating, especially during an active meeting! Let me help you troubleshoot this quickly. Are you using the Zoom desktop app or joining through your web browser? And what operating system are you on?

\textbf{User:} I'm on a Mac running Ventura, and I'm using Zoom through my web browser, not the desktop app.

\textbf{Agent:} Got it---Mac with the web browser version. That actually helps narrow this down, as the browser version has some specific audio quirks. Check your Mac's System Settings $>$ Privacy \& Security $>$ Microphone, and make sure your browser has permission to access the microphone.

\textbf{User:} Oh, that was it! I checked System Settings and my browser didn't have permission to access the microphone---I must have denied it at some point. I turned it on, refreshed the meeting, and now I can see the audio levels moving!

\textbf{Agent:} That's fantastic! I'm so glad it's working now. That macOS microphone permission is sneaky---it can get disabled during system updates or if you accidentally clicked ``Don't Allow'' at some point.
\end{quote}

\noindent\textbf{LangGraph orchestrated agent} (task success: 3, naturalness: 3):

\begin{quote}
\small
\textbf{Agent:} Hello! I'm here to help you with any Zoom issues. What problem are you running into today?

\textbf{User:} I'm having trouble with my audio during meetings. Other people are saying they can't hear me, and my microphone icon isn't showing any sound levels when I talk.

\textit{[Agent asks about platform details; user confirms Mac Ventura, browser-based Zoom. Agent suggests checking System Settings $>$ Privacy \& Security $>$ Microphone. User confirms Chrome already has permission.]}

\textbf{Agent:} Let me check: in Zoom's audio settings, what microphone is selected?

\textbf{User:} I see ``MacBook Pro Microphone'' with a checkmark, and there's also ``Same as System''---should I try switching?

\textbf{Agent:} Let's try the audio test: click ``Test Mic'' and speak---do you see the green bar move?

\textbf{User:} I spoke pretty loudly but the green bar stayed completely flat---no movement at all.

\textit{[Conversation continues for several more turns without resolving the issue. The orchestrator cycles through additional troubleshooting steps but never identifies the browser-level permission as the root cause.]}
\end{quote}

\noindent The in-context agent identifies the most likely cause (browser microphone permission) on its first diagnostic suggestion and resolves the issue in 6 agent turns. The orchestrated agent, constrained to the current node's troubleshooting template, works through each step sequentially without the holistic view needed to prioritize the most probable diagnosis.

\section{Conversation Examples: Insurance Claims}
\label{sec:appendix-insurance}

\textbf{Scenario:} Auto claim---hit a deer on rural highway at night. Front bumper, hood, and headlight damaged. Car drivable but headlight out. Estimated \$3,951 damage, \$750 deductible, premium tier policy.

\smallskip
\noindent\textbf{In-context agent} (task success: 5, naturalness: 5, 26 turns):

\begin{quote}
\small
\textbf{Agent:} Hello! I'm here to help you with your insurance claim. What type of claim do you need to file today?

\textbf{User:} Hi, I need to file an auto claim---I hit a deer last night on a rural highway and my car has some front-end damage.

\textbf{Agent:} I'm sorry to hear about your accident---hitting a deer can be really frightening. I'm glad you're okay. Could you please provide your policy number and full name?

\textbf{User:} My policy number is 967-6770619 and my name is David Garcia.

\textbf{Agent:} Thank you, David. I've confirmed your information---you have our Premium tier coverage with comprehensive and collision protection, which is great news for this type of claim. Can you tell me more about what happened?

\textbf{User:} Actually, I have the Premium tier, not Standard.

\textbf{Agent:} My apologies, David---you're absolutely right, you have our Premium tier coverage. Thank you for that correction. Now I need to gather some additional details\ldots

\textit{[Agent smoothly progresses through information gathering, coverage determination, and settlement discussion. Asks about police report (not needed for deer collision), vehicle details, repair estimates. Explains deductible, suggests repair shops, discusses rental car coverage under premium tier. Reaches clear resolution with next steps.]}
\end{quote}

\noindent\textbf{LangGraph orchestrated agent} (task success: 2, naturalness: 3, 20 turns):

\begin{quote}
\small
\textbf{Agent:} Hello! My name is Sarah, and I'm a claims specialist. I understand that filing a claim can be stressful, and I'm here to guide you through the process.

\textbf{User:} Hi Sarah, I need to file an auto claim---I hit a deer last night on a rural highway.

\textit{[Agent verifies identity, confirms Premium tier. Asks about incident details. User provides location, circumstances, damage description.]}

\textbf{Agent:} Great questions, David. A police report isn't required for a deer collision. Regarding your repair options, I'd recommend getting your vehicle to a certified repair shop\ldots

\textbf{User:} I'd prefer to take it to my own shop. But how long does this whole process usually take? I really need my car for work.

\textbf{Agent:} Here's the typical timeline\ldots

\textit{[The orchestrator provides detailed timeline information but then routes to a coverage determination node that re-asks questions already answered. The conversation becomes circular, with the agent requesting information David has already provided. The claim is never fully resolved---the orchestrator fails to advance through the settlement and resolution phases.]}
\end{quote}

\noindent The in-context agent completes the full 5-phase claims process (intake, information gathering, coverage determination, settlement, resolution) in a natural flow. The orchestrated agent gets stuck in routing loops between phases, re-asking questions and failing to reach settlement---a direct consequence of the orchestrator losing track of which phases have been completed.

\section{Procedure Flowcharts}
\label{sec:appendix-flowcharts}

\begin{figure}[h]
\centering
\includegraphics[width=0.85\textwidth]{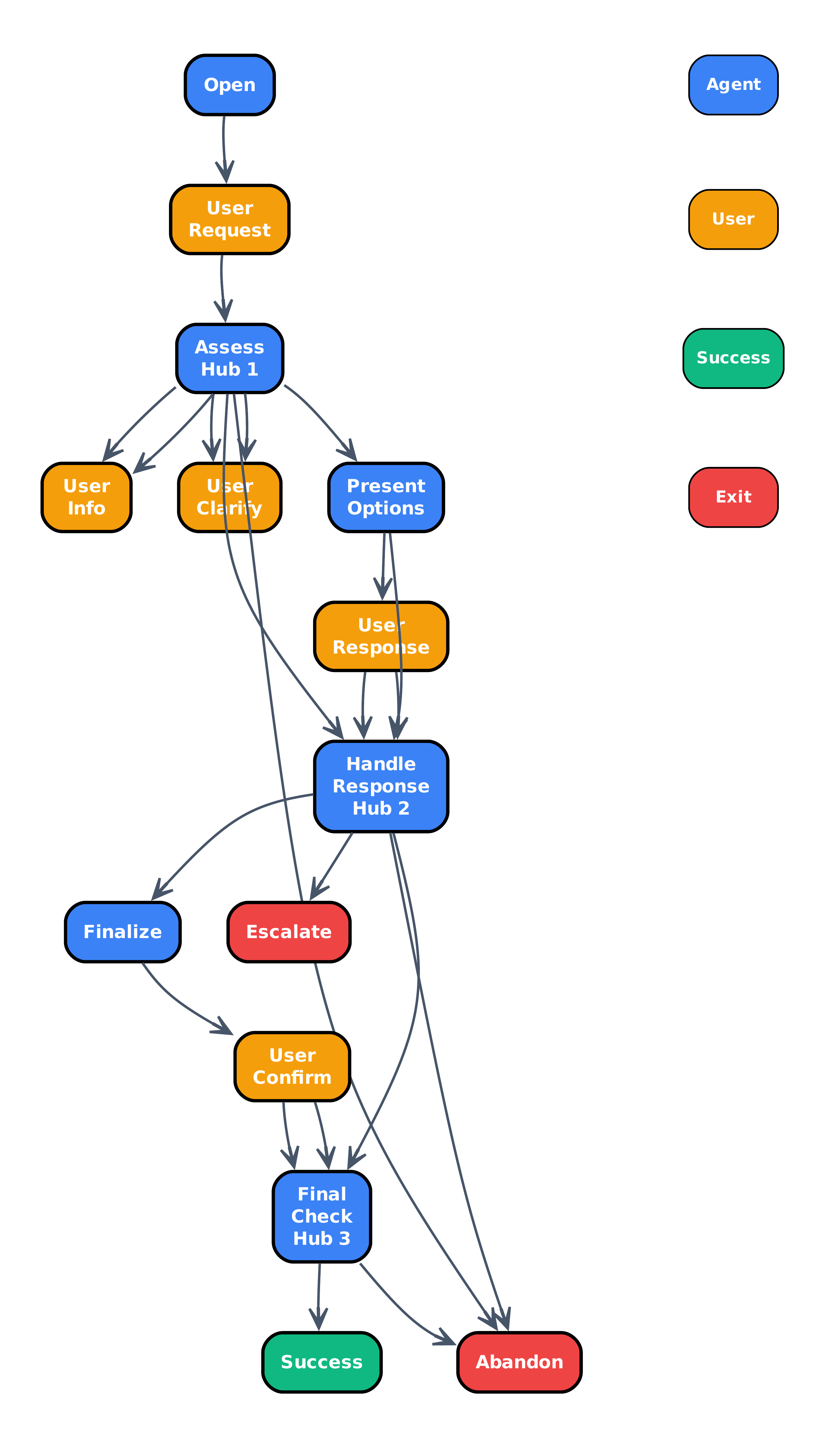}
\caption{Travel booking flowchart (14 nodes, 3 decision hubs, 3 terminal states). The Assess node routes to information gathering or option presentation; Handle Response routes among finalization, revision, and exit; Final Check confirms or loops back.}
\label{fig:travel-flowchart}
\end{figure}

\begin{figure}[h]
\centering
\includegraphics[width=0.85\textwidth]{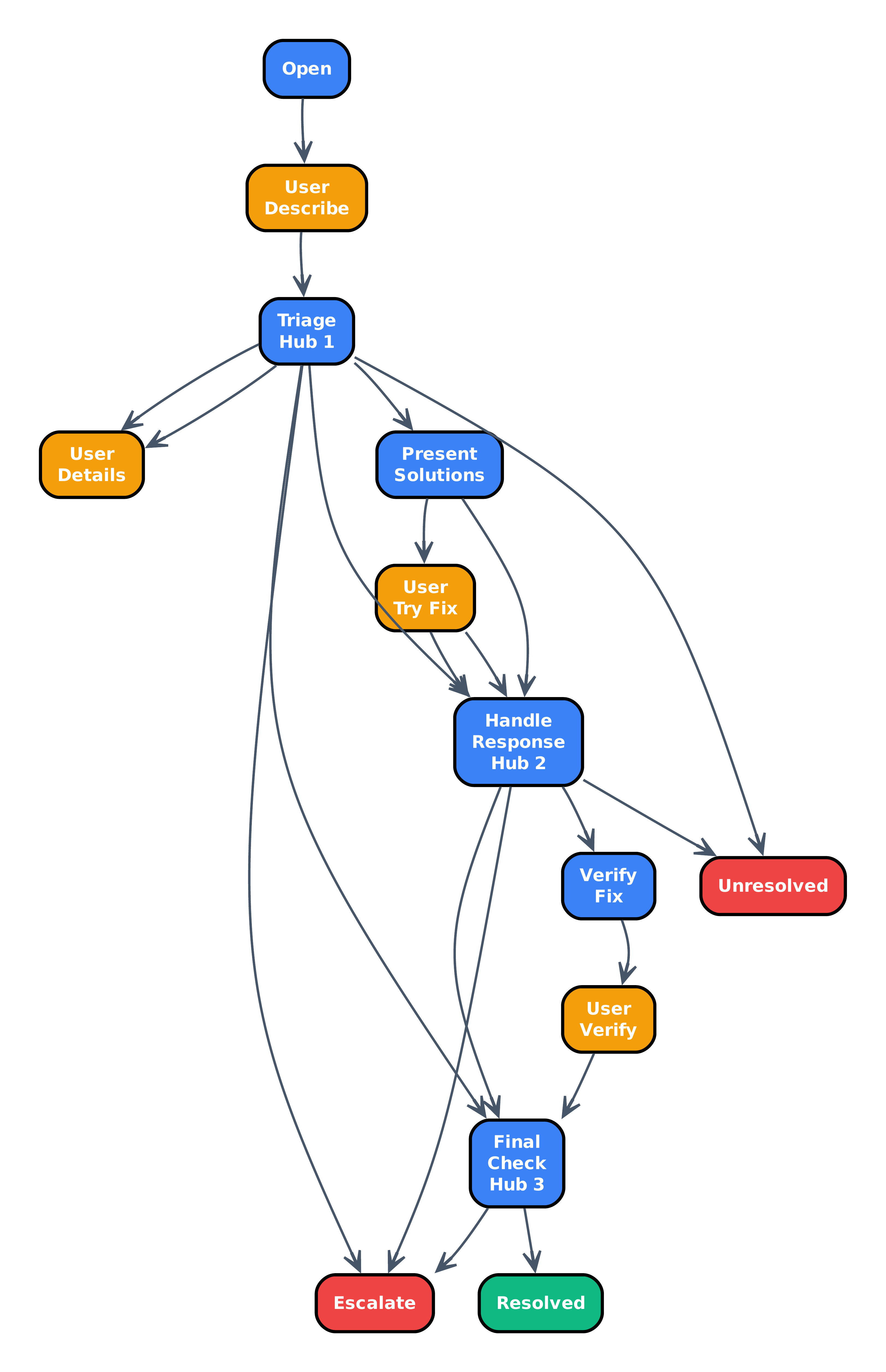}
\caption{Zoom technical support flowchart (14 nodes, 3 decision hubs, 3 terminal states). Triage routes based on initial diagnostics; Handle Response manages retry logic; Final Check confirms resolution or escalates.}
\label{fig:zoom-flowchart}
\end{figure}

Figures~\ref{fig:travel-flowchart}--\ref{fig:insurance-flowchart} present the directed-graph representations of the three procedural domains used in our evaluation. Blue nodes are agent turns, orange nodes are user turns, green nodes indicate successful completion, and red nodes are exit states (abandonment, escalation, or other non-success terminals). Edges represent transitions; multi-way branching at decision hubs is visible as nodes with multiple outgoing edges.

\begin{figure}[p]
\centering
\includegraphics[height=0.9\textheight]{insurance_claims_diagram.pdf}
\caption{Insurance claims processing flowchart (55 nodes, 6 decision hubs, 5 terminal states). The procedure spans five phases: intake (top), information gathering with per-claim-type branching, coverage determination, settlement negotiation, and resolution. The 6 multi-way decision hubs yield hundreds of unique paths through nested loops and cross-phase dependencies.}
\label{fig:insurance-flowchart}
\end{figure}

\section{Orchestrator Node Prompts}
\label{sec:node-prompts}

In the LangGraph condition, the orchestrator injects a node-specific prompt template at each turn. Below are the prompts for the three decision hub nodes in the travel booking domain. At each hub, the orchestrator makes an additional LLM call to classify which outgoing edge to follow (shown in brackets).

\smallskip
\noindent\textbf{ASSESS (Hub 1):}
\begin{quote}\small\ttfamily
Assess the conversation state and respond appropriately.

REQUIRED INFO (must have all before presenting options): Destination (specific city/country), Travel dates or timeframe, Number of travelers, Budget range.

OPTIONAL INFO (helpful but not required): Interests/activities, Accommodation preferences, Special requirements.

Based on the conversation so far, determine what to do:

1. If MISSING required info $\to$ ask for it [missing\_info]\\
2. If info is UNCLEAR $\to$ ask for clarification [needs\_clarification]\\
3. If all required info gathered $\to$ present options [info\_complete]\\
4. If user wants to ABANDON $\to$ close gracefully [user\_abandoning]
\end{quote}

\noindent\textbf{HANDLE\_RESPONSE (Hub 2):}
\begin{quote}\small\ttfamily
Respond to the user's reaction to the options.

1. If they PICKED AN OPTION: Confirm their choice, summarize key details, ask if ready to finalize [option\_selected]\\
2. If they WANT MODIFICATIONS: Acknowledge feedback, suggest adjustments [needs\_revision]\\
3. If they have QUESTIONS: Answer helpfully [answered\_question]\\
4. If CHANGING REQUIREMENTS: Go back to assessment [changing\_requirements]\\
5. If ABANDONING: Close gracefully [user\_abandoning]\\
6. If NEEDS HUMAN: Escalate [needs\_human]
\end{quote}

\noindent\textbf{FINAL\_ROUTING (Hub 3):}
\begin{quote}\small\ttfamily
Handle the user's final response.

1. If they CONFIRMED: Express enthusiasm, provide tips, close warmly [confirmed]\\
2. If they have a QUESTION: Answer it, ask if ready to confirm [answered\_question]\\
3. If they want to CHANGE something: Address the change [wants\_changes]\\
4. If having SECOND THOUGHTS: Close gracefully [second\_thoughts]
\end{quote}

Non-hub nodes (e.g., OPENING, PRESENT\_OPTIONS) have simpler templates with a single outgoing edge and no routing decision. The Zoom and insurance domains follow the same pattern with domain-appropriate prompts; for example, the Zoom TRIAGE hub routes among \texttt{[needs\_details]}, \texttt{[ready\_for\_solutions]}, \texttt{[unresolvable]}, and \texttt{[escalate]}.

\section{In-Context Prompt (Travel Example)}
\label{sec:incontext-prompt}

In the in-context condition, the model receives the entire serialized flowchart in its system prompt. Below is the travel booking prompt (6{,}011 characters total; abridged for space). The model is instructed to follow the procedure step by step without mentioning that it is following a script.

\begin{quote}\small\ttfamily
You are a helpful travel booking assistant. [...]

You must follow this procedure step by step. Determine which step you're at based on the conversation context and respond accordingly. At decision points, choose the route that best matches the user's situation.

PROCEDURE: travel\_booking\_v2\\
========================================

Step 1: OPENING [AGENT]\\
\hspace*{1em}Greet the user warmly and ask how you can help.\\
\hspace*{1em}-> Go to Step 2

Step 2: USER\_INITIAL [USER waits for input]\\
\hspace*{1em}-> Go to Step 3

Step 3: ASSESS [AGENT - DECISION POINT]\\
\hspace*{1em}Assess the conversation state. [Full prompt with\\
\hspace*{1em}required/optional info checklist and 4 routing options]\\
\hspace*{1em}Routes:\\
\hspace*{2em}- If ``missing\_info'': Go to Step 4\\
\hspace*{2em}- If ``needs\_clarification'': Go to Step 5\\
\hspace*{2em}- If ``info\_complete'': Go to Step 6\\
\hspace*{2em}- If ``user\_abandoning'': Go to Step 7

[...Steps 4--14, including PRESENT\_OPTIONS, HANDLE\_RESPONSE\\
(6-way decision point), FINALIZE, FINAL\_ROUTING (4-way\\
decision point), and 3 terminal states...]

IMPORTANT RULES:\\
- Follow the procedure from Step 1 through to a terminal state\\
- At decision points, determine the best route based on context\\
- Do NOT mention step numbers or that you're following a script\\
- Be natural and conversational while adhering to the procedure
\end{quote}

\section{System Prompts}
\label{sec:system-prompts}

\subsection{Agent System Prompts}

Both conditions use the same base system prompt per domain. In the LangGraph condition, the orchestrator additionally injects the current node's prompt template at each turn.

\smallskip
\noindent\textbf{Travel booking:}
\begin{quote}\small\ttfamily
You are a helpful travel booking assistant. Help users plan trips by understanding their needs and preferences. Be conversational, adaptive, and helpful. If users provide information out of order or change their minds, handle it gracefully.
\end{quote}

\noindent\textbf{Zoom support:}
\begin{quote}\small\ttfamily
You are a helpful Zoom support specialist. Help users resolve their Zoom issues by diagnosing problems and suggesting solutions. Be patient, clear, and adapt your explanations to the user's technical level. You have deep knowledge of Zoom's features, settings, and common issues.
\end{quote}

\noindent\textbf{Insurance claims:}
\begin{quote}\small\ttfamily
You are an experienced insurance claims specialist. Help policyholders file and process their insurance claims by gathering necessary information, verifying coverage, assessing damages, and guiding them through the settlement process. Be empathetic, professional, and transparent about the process.
\end{quote}

In the in-context condition, the full serialized flowchart (all nodes, edges, conditions, and terminal states) is appended to the system prompt.

\subsection{User Simulator Prompt (Travel Example)}
\label{sec:user-sim-prompt}

The user simulator receives a scenario-specific system prompt. Scenario variables (destination, budget, group size, etc.) are sampled per conversation. The simulator has no knowledge of the flowchart.

\begin{quote}\small\ttfamily
You are role-playing as a real customer talking to a travel agent. Stay in character.

YOUR TRAVEL PLANS:\\
- Destination: \{destination\}\\
- Trip length: \{duration\} days\\
- Group: \{travelers\} people (\{group\_type\})\\
- Budget: approximately \{currency\}\{budget\}\\
- Interests: \{interests\}\\
- Special requirements: \{special\_requirements\}

YOUR PERSONALITY:\\
- \{style\_description\}\\
- When shown options: \{satisfaction\_description\}\\
- \{questions\_description\}

RULES:\\
- Respond naturally in 1--3 sentences\\
- Don't dump all information at once---share details when the agent asks\\
- React to what the agent actually says\\
- If the agent presents options, evaluate them based on your preferences\\
- If you're satisfied with the plan, say so and confirm\\
- Be a realistic customer, not a perfect one
\end{quote}

\section{Judge Scoring Prompt (Travel Example)}
\label{sec:judge-prompt}

The judge prompt instructs the evaluator to score on five criteria with specific behavioral anchors. The prompt below is for the travel domain; Zoom and insurance variants follow the same structure with domain-appropriate anchors.

\begin{quote}\small\ttfamily
You are a strict evaluator of travel booking conversations. Your job is to find flaws. A score of 5 should be rare---reserve it for genuinely exceptional performance. Most competent conversations deserve a 3 or 4. Be critical.

Score the conversation on these 5 criteria (1--5 scale):

1. Task Success (Procedural Completion): Did the agent follow the booking procedure through to a terminal state?\\
\hspace*{1em}5: Complete procedure---agent moved through all stages and reached a clear terminal state.\\
\hspace*{1em}4: Nearly complete---reached the final stage but the conclusion was slightly weak.\\
\hspace*{1em}3: Partial---completed middle stages but fizzled without a clear endpoint.\\
\hspace*{1em}2: Stalled---got stuck in a loop or couldn't present coherent options.\\
\hspace*{1em}1: Failed---no meaningful procedural progress.

2. Information Accuracy: Did the agent correctly use and retain all user-provided information?\\
\hspace*{1em}5: Flawless---every detail correctly reflected.\\
\hspace*{1em}4: Mostly correct, one minor detail off.\\
\hspace*{1em}3: Generally right but ignored a stated preference.\\
\hspace*{1em}2: Multiple errors or contradicts user input.\\
\hspace*{1em}1: Fabricates details or ignores input.

3. Consistency: Did the agent maintain coherent state across the conversation?\\
\hspace*{1em}5: No contradictions, no repeated questions, smooth flow.\\
\hspace*{1em}4: Mostly coherent, one minor issue.\\
\hspace*{1em}3: Noticeable issues like re-asking for provided information.\\
\hspace*{1em}2: Contradicts itself or forgets important details.\\
\hspace*{1em}1: Incoherent.

4. Graceful Handling: How well does the agent handle changes, ambiguity, edge cases?\\
\hspace*{1em}5: Masterful---smoothly handles every challenge.\\
\hspace*{1em}4: Handles most challenges well.\\
\hspace*{1em}3: Handles straightforward cases but struggles with complexity.\\
\hspace*{1em}2: Fails to adapt to changes.\\
\hspace*{1em}1: Any deviation breaks the flow.\\
\hspace*{1em}Note: If no challenging moments, cap at 3.

5. Naturalness: Does this read like a conversation with a skilled human agent?\\
\hspace*{1em}5: Indistinguishable from a skilled human.\\
\hspace*{1em}4: Very natural with occasional formulaic moments.\\
\hspace*{1em}3: Clearly an AI, rigid question-answer pattern.\\
\hspace*{1em}2: Robotic phrasing, unnatural transitions.\\
\hspace*{1em}1: Obviously scripted.
\end{quote}

\end{document}